\pdfoutput=1

\documentclass[11pt]{article}

\usepackage[]{ACL2023}

\usepackage{times}
\usepackage{latexsym}

\usepackage[T1]{fontenc}

\usepackage[utf8]{inputenc}

\usepackage{microtype}

\usepackage{multirow}
\usepackage{multicol}
\usepackage{xspace}
\usepackage{colortbl}
\usepackage{makecell}
\usepackage{graphicx}
\usepackage{booktabs}
\usepackage{amsmath}
\usepackage{arydshln}
\usepackage{amssymb}
\usepackage{tikz}
\usepackage{pgfplots}
\pgfplotsset{compat=1.8}
\usepgfplotslibrary{statistics}
\usepackage{makecell}
\usepackage{subcaption}

\usepackage{inconsolata}
\usepackage{hyperref}
\usepackage{amsfonts}
\usepackage[most]{tcolorbox}
\usepackage{dashrule}
\usepackage{ragged2e} 

\usepackage{inconsolata}

\newcommand\benchall{\textsc{TritonBench}\xspace}
\newcommand\benchone{\textsc{TritonBench-{G}}\xspace}
\newcommand\benchtwo{\textsc{TritonBench-{T}}\xspace}

\title{\benchall: Benchmarking Large Language Model Capabilities for Generating Triton Operators}

\author{
    \textbf{Jianling Li\textsuperscript{1,}\footnotemark[1]},
    \textbf{Shangzhan Li\textsuperscript{2,}\footnotemark[1]},
    \textbf{Zhenye Gao\textsuperscript{3}},
    \textbf{Qi Shi\textsuperscript{4,}\footnotemark[2]},
    \textbf{Yuxuan Li\textsuperscript{4,}\footnotemark[2]}, 
    \textbf{Zefan Wang\textsuperscript{4}},
    \\
    \textbf{Jiacheng Huang\textsuperscript{4}}, 
    \textbf{Haojie Wang\textsuperscript{4}}, 
    \textbf{Jianrong Wang\textsuperscript{1}}, 
    \textbf{Xu Han\textsuperscript{4}}, 
    \textbf{Zhiyuan Liu\textsuperscript{4}}, 
    \textbf{Maosong Sun\textsuperscript{4}}
    \\
\\    
\textbf{\textsuperscript{1}} Tianjin University, Tianjin, China
\\
\textbf{\textsuperscript{2}} Harbin Institute of Technology, Harbin, China 
\\
\textbf{\textsuperscript{3}} The Hong Kong University of Science and Technology (Guangzhou), Guangzhou, China
\\
\textbf{\textsuperscript{4}} Tsinghua University, Beijing, China
\\
}

\begin{document}
\maketitle

\renewcommand{\thefootnote}{\fnsymbol{footnote}}
\footnotetext[1]{Equal contribution.}
\footnotetext[2]{Corresponding authors.}
\renewcommand{\thefootnote}{\arabic{footnote}}

\begin{abstract}

Triton, a high-level Python-like language designed for building efficient GPU kernels, is widely adopted in deep learning frameworks due to its portability, flexibility, and accessibility.
However, programming and parallel optimization still require considerable trial and error from Triton developers. 
Despite advances in large language models (LLMs) for conventional code generation, these models struggle to generate accurate, performance-optimized Triton code, as they lack awareness of its specifications and the complexities of GPU programming. 
More critically, there is an urgent need for systematic evaluations tailored to Triton.
In this work, we introduce \benchall, the first comprehensive benchmark for Triton operator generation. 
\benchall features two evaluation channels: a curated set of $184$ real-world operators from GitHub and a collection of operators aligned with PyTorch interfaces. 
Unlike conventional code benchmarks prioritizing functional correctness, \benchall also profiles efficiency performance on widely deployed GPUs aligned with industry applications. 
Our study reveals that current state-of-the-art code LLMs struggle to generate efficient Triton operators, highlighting a significant gap in high-performance code generation. 
\benchall will be available at \url{https://github.com/thunlp/TritonBench}.
\end{abstract}

\section{Introduction}
Triton~\cite{tillet2019triton} language, a high-level Python-like programming language designed for implementing efficient GPU kernels, is playing an increasingly pivotal role in the ever-scaling deep learning ecosystems~\cite{abadi2016tensorflow,paszke2019pytorch}. 
Due to the superior portability, flexibility, lightweight design, and accessibility to less proficient programmers, Triton is prevalently adopted in modern Large Language Model (LLM) frameworks such as vLLM~\cite{vllm}, LightLLM~\cite{lightllm}, Liger-kernel~\cite{liger} and unsloth~\cite{unsloth}. 
However, crafting high-performance operators remains challenging, especially for the intricate balance between memory hierarchy management, parallel thread coordination, and hardware-specific optimizations. 
Even though Triton abstracts away many complexities of low-level programming architectures like CUDA, it still requires developers to manually handle critical aspects such as pointer arithmetic and memory access patterns, making performance tuning a labor-intensive process that often involves extensive trial and error.

Current research in AI-assisted coding has reached a human-competitive level~\cite{hui2024qwen2,zhu2024deepseek}, yet it is primarily restricted to general-purpose languages like Python. 
However, LLMs still face challenges in generating Domain Specific Language (DSL) code. 
Specifically for Triton, current models might be unfamiliar with Triton specification and the intricacies of GPU programming~\cite{nichols2024can}. 
Most importantly, the ability of these models to produce high-quality Triton code remains unassessed. 
Therefore, a high-quality benchmark paired with performance-aware metrics is urgently required.

\begin{figure*}[t]
\vspace{-5mm}
    \centering
    \includegraphics[width=0.98 \textwidth]{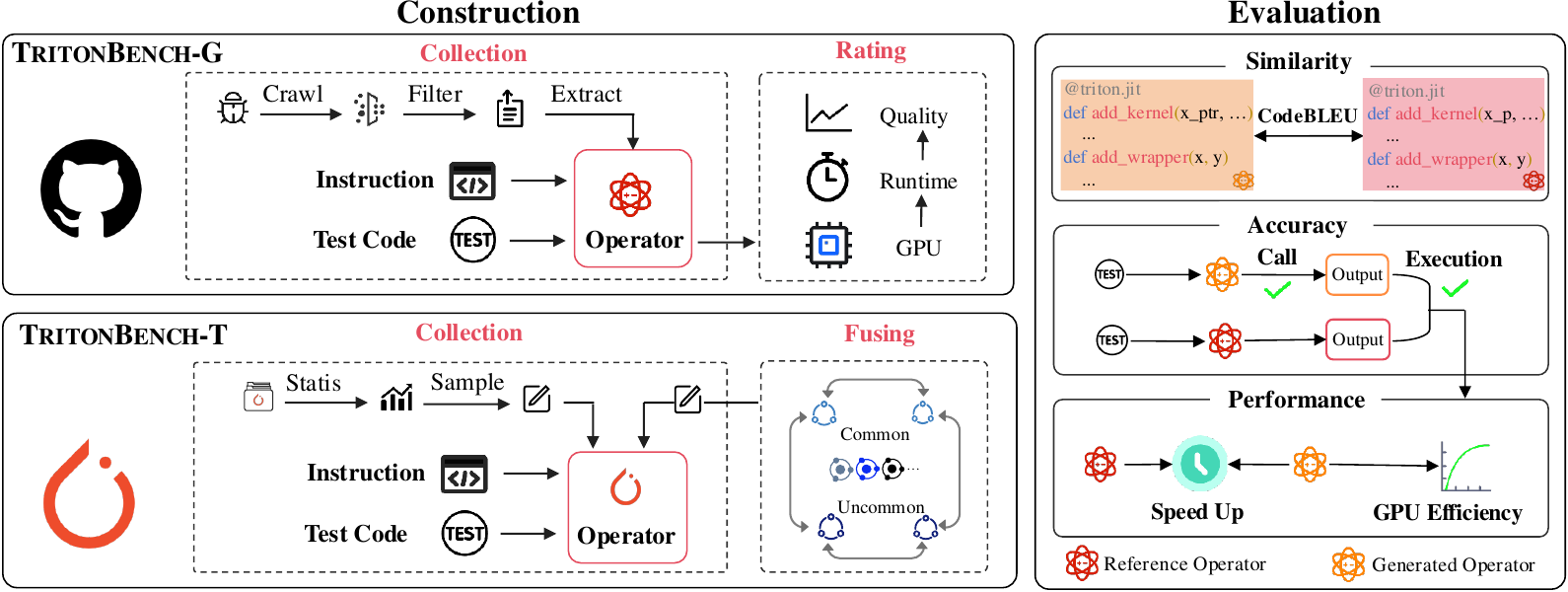}
    \caption{Illustration of the construction and evaluation of \benchall.}
    \label{fig:main_figure}
\end{figure*}

In this study, we present \benchall, a performance-aware benchmark framework for Triton generation, which contains two channels, namely \benchone and \benchtwo. 
Specifically, \benchone contains $184$ carefully curated operators from existing GitHub repositories, reflecting the realistic demand for Triton operator development. 
As a complement, \benchtwo is composed of operator development tasks aligned with PyTorch interfaces, covering operators under-represented by public sources.
Moreover, unlike the majority of code benchmarks merely prioritizing functional correctness~\cite{chen2021evaluating,austin2021program}, \benchall emphasizes efficiency performance profiling against reference programs on NVIDIA GPUs, better aligning industrial demands. 

As shown in Figure~\ref{fig:main_figure}, for \benchone, we follow three steps: 1) scrape and collect high-quality operators, 2) generate instructions via prompts, and 3) annotate test code with LLMs.
Moreover, HPC experts evaluate GPU performance for all triton codes. 
For \benchtwo, we provide operator generation tasks aligned with PyTorch.
To construct these tasks, we first perform a frequency analysis to select torch operators, combine them into diverse sets, and provide paired instructions and test code. 
Our evaluation metrics include similarity, call and execution accuracy, speed up, and GPU efficiency.

We conduct extensive experiments across a broad range of LLMs. Overall, the difficulty of \benchone is greater than that of \benchtwo. 
The highest execution accuracy on \benchone can reach $23.91$\%, while on \benchtwo, it can reach $53.01$\%. 
For all correctly executed operators generated by the models, the best speed up on \benchone is $1.56\times$, whereas, on \benchtwo, it is $1.91\times$. 
Additionally, we perform in-depth analyses of LLMs’ behavior on \benchall and summarize the challenges in Triton generation. 
The results reveal that current LLMs are not yet fully capable of handling \benchall, underscoring the challenge of enabling LLMs to generate Triton code effectively. 
We hope this work initiates evaluation in this under-explored area and fosters advancements in LLM-driven operator development.
\section{Related Work}

\subsection{Triton Development}
Triton~\cite{triton-2021-translation} is an open-source, Python-like language and a compiler designed to simplify GPU programming in AI and HPC. 
It abstracts the complexities of CUDA by introducing a block-based programming model, automating low-level optimizations such as memory coalescing and tensor core utilization, and making it more accessible to researchers without HPC background. 
Nonetheless, Triton provides explicit control over memory access patterns and parallelism. 
This balance of productivity and flexibility makes it prevalently adopted in both academia and industry~\cite{vllm, lightllm, liger, unsloth}. 
However, Triton developers must still laboriously tune critical parameters to exploit hardware capabilities. 
LLM code generation poses prospects for automating Triton development, which calls for a systematic evaluation of generated operators.

\subsection{Code Benchmarks}
The demand for proper measurement of coding capability arises as the program synthesis research advances. 
The primary practice of coding benchmarks is functional correctness testing, usually realized by test case construction and sandbox execution. 
For example, \textsc{HumanEval}~\cite{humeval-2021-human} curate hand-written programs and test cases, and MBPP~\cite{mbpp2021} create programming problems by crowd-sourcing. 
The functionality test has recently extended to automated test generation for better coverage~\cite{evalplus} and broader applications, including software engineering~\cite{jimenez2024swebench}.
Another vital aspect of coding benchmarking is performance profiling~\cite{shypulalearning,evalplus,huang2024effibench,qiu2024efficient}.
However, most existing frameworks focus on competition-style, single-process execution. 
While there are some frameworks for evaluating parallel programming on CPUs~\cite{nichols2024can,chaturvedi2024hpc}, benchmarks targeting GPU code remain scarce. 
As the deployment of deep learning models scales up, a comprehensive evaluation framework that considers both correctness and performance on GPU code becomes increasingly necessary.

\subsection{LLMs for Code Generation}
LLMs have recently demonstrated impressive capabilities in generating code from natural language instructions, as evidenced by models such as DeepSeek-Coder~\cite{guo2024deepseek,zhu2024deepseek} and Qwen-Coder~\cite{hui2024qwen2}, which have achieved strong performance on broad coding benchmarks. 
Despite their versatility, they often struggle with Domain-Specific Languages~(DSLs) designed for higher levels of abstraction and improved efficiency in targeted contexts~\cite{wkasowski2023domain}. 
The main reason for this status is the limited availability of DSL datasets and benchmarks~\cite{cassano2024knowledge, pujar2023automated}, coupled with their unique syntax and semantics~\cite{pujar2023automated}, posing significant challenges for LLMs~\cite{buscemi2023comparative}.
In this work, we focus on DSLs within the high-performance computing domain where the challenges we mentioned are more pronounced for involving the parallel programming model. 
We introduce the first comprehensive benchmark for Triton generation, providing a systematic evaluation framework that aims to guide future improvements in DSL-centric LLM code generation.
\section{\benchone}
\label{sec:git}

\begin{figure}[ht]
    \centering
    \includegraphics[width=0.48\textwidth]{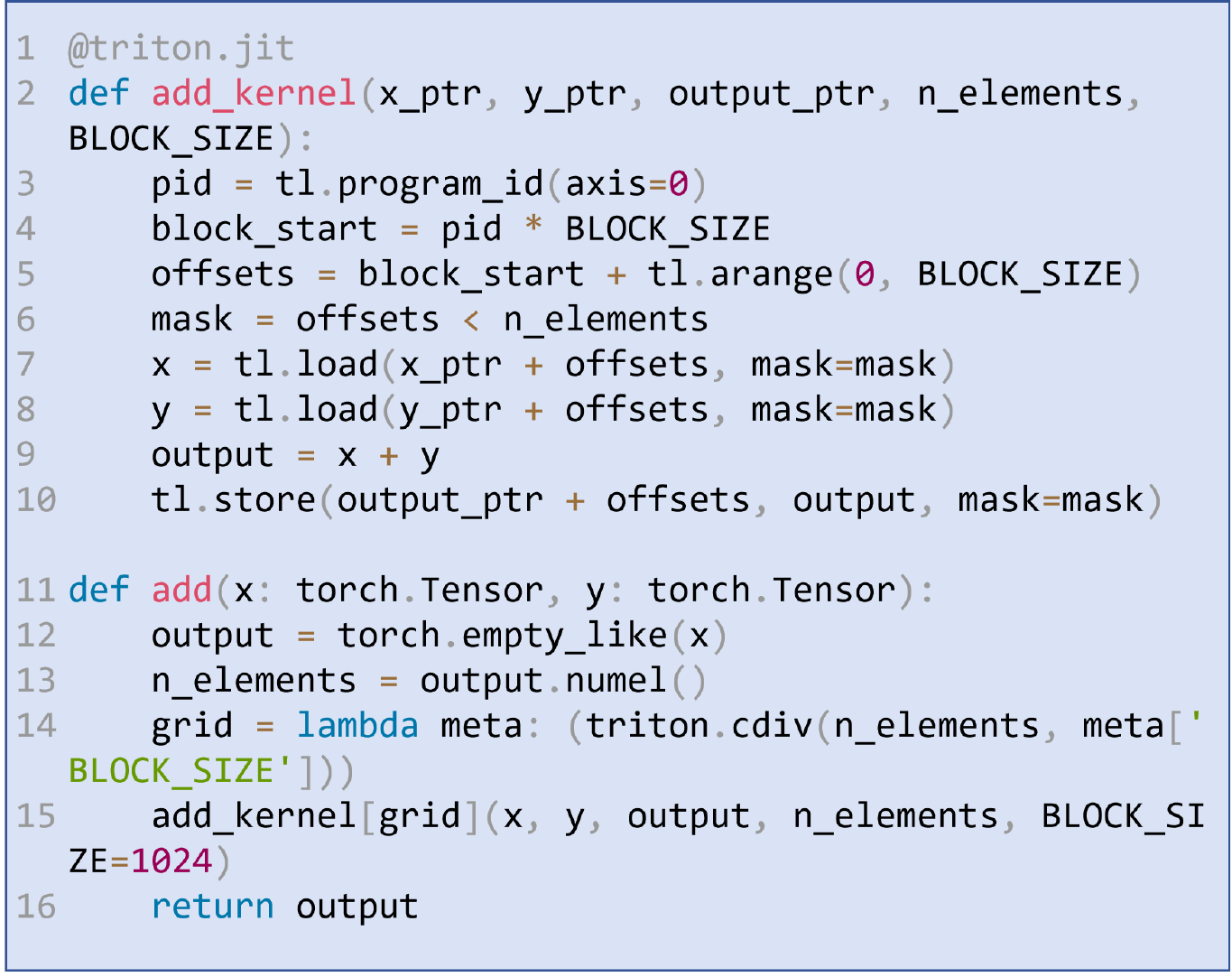}
    \caption{
    Implementation of the Triton ``add'' operator. Lines $3$-$6$ perform for tensor element addressing, followed by the calculation and storage in lines $7$-$10$. 
    The kernel is called in wrapper line $15$.
    }
    \label{fig:triton_example}
    \vspace{-3mm}
\end{figure}

Triton~\cite{tillet2019triton} is a DSL that abstracts away low-level complexities to simplify GPU programming for computation-intensive tasks, with flexibility for specialized applications like machine learning.
Typically, a Triton operator includes at least a kernel and a wrapper. 
The kernel comprises code executed on the GPU, focusing on tensor element addressing and thread parallel coordination. 
Meanwhile, the wrapper offers a Python function that encapsulates the kernel call.
Figure \ref{fig:triton_example} shows an example of Triton operator. 

We create \benchone by curating high-quality human-authored Triton operators from Github, which reflects Triton’s currently actual requirements. 
The following sections will explain data collection (\S~\ref{sec:benchone-datacollect}), data statistics (\S~\ref{sec:benchone-datastatistic}), operator quality rating (\S~\ref{sec:benchone-qualityrating}), test code design (\S~\ref{sec:benchone-testcodedesign}), and evaluation metrics (\S~\ref{sec:benchone-evalmetric}).

\subsection{Data Collection}
\label{sec:benchone-datacollect}
Our process starts by gathering Triton-related GitHub repositories with more than $100$ stars, which collectively encompass $95$ repositories with $845$ Python files. 
As Triton repositories with higher star counts are rare, $100$ stars serve as an optimal threshold, striking a balance between quality and quantity.
We then use prompt-based filtering (see prompt~\ref{prompt_filter} in the Appendix) to process the candidate Python files and select $250$ that specifically contain Triton code snippets.

Afterward, we perform a rigorous manual inspection of the Triton code to ensure its accuracy and clarity. 
This process involves filling in missing components, removing redundant sections, and debugging the operators.
When a file contains multiple independent Triton operators, we split them into separate files. For operators that are solely kernels, we add the necessary wrappers to ensure they work as intended. Additionally, to ensure uniqueness, we leverage \textsc{CodeBertScore}~\cite{codebertscore2023} to eliminate duplicates.

Finally, we generate the LLM instruction for each operator based on prompt~\ref{prompt_instru}. 
The instructions provide essential details, including the operator’s functionality, corresponding function names, and a comprehensive input/output demonstration. 
All instructions are carefully reviewed and manually verified to ensure they correctly reflect the intended behavior of each operator.

\subsection{Data Statistics}
\label{sec:benchone-datastatistic}
Table~\ref{tab:statis_git} summarizes statistics of \benchone. 
In this benchmark, each operator is assigned a difficulty level, from $\bf d1$ (easiest) to $\bf d5$ (most challenging), by an LLM guided by prompt~\ref{prompt_diff}, with subsequent manual verification by two domain experts. 
For each difficulty level, we report statistics including the average number of functions~(\textbf{func\#}), parameters~(\textbf{params\#}), lines~(\textbf{lines\#}), and tokens~(\textbf{tok\#}). 
Notably, the upward trend observed in these statistics as the difficulty level increases suggests the expert-driven grading scheme is largely reasonable. 

Compared to existing code generation tasks \cite{chen2021evaluating, austin2021program}, the average instruction length in \benchone is substantially longer, which is a deliberate design decision. 
The extended instructions provide richer context, which can help the model understand nuanced requirements and generate high-quality operators. 
Additionally, this approach better reflects real-world operator development practices where detailed requirements are indispensable. 

\begin{table}[t!]
    \centering
    \resizebox{0.49\textwidth}{!}{
    \begin{tabular}{lcccccc}
    \toprule
    \multirow{2}{*}{\textbf{Difficulty}}
     & \multicolumn{1}{c}{\textbf{Instruction}} & \multicolumn{5}{c}{\textbf{Triton Operator}}                           \\ \cmidrule(lr){2-2} \cmidrule(lr){3-7} 
                            & \textbf{tok\#}    &\textbf{func\#} & \textbf{params\#} & \textbf{line\#} & \textbf{tok\#}   \\ \midrule
    $\mathbf{d1}$ ($1.6$\%)   & $296.67$  & $2.00$  & $1.33$  & $26.00$      & $369.0$         \\
    $\mathbf{d2}$ ($14.7$\%)  & $363.26$  & $2.41$  & $2.70$  & $45.56$      & $678.1$         \\ 
    $\mathbf{d3}$ ($35.3$\%)  & $353.80$  & $3.80$  & $3.34$  & $102.42$     & $1510.4$        \\ 
    $\mathbf{d4}$ ($45.7$\%)  & $394.48$  & $3.89$  & $6.04$  & $153.77$     & $2689.1$        \\ 
    $\mathbf{d5}$ ($2.7$\%)   & $469.60$  & $6.60$  & $6.00$  & $249.80$     & $4581.4$        \\ 
    
    \bottomrule
    \end{tabular}
       }
    \caption{Statistics of \benchone.}
    \label{tab:statis_git}
\end{table}

\subsection{Operators Quality Rating}
\label{sec:benchone-qualityrating}
\begin{figure}[t!]
    \centering
    \includegraphics[width=0.28\textwidth]{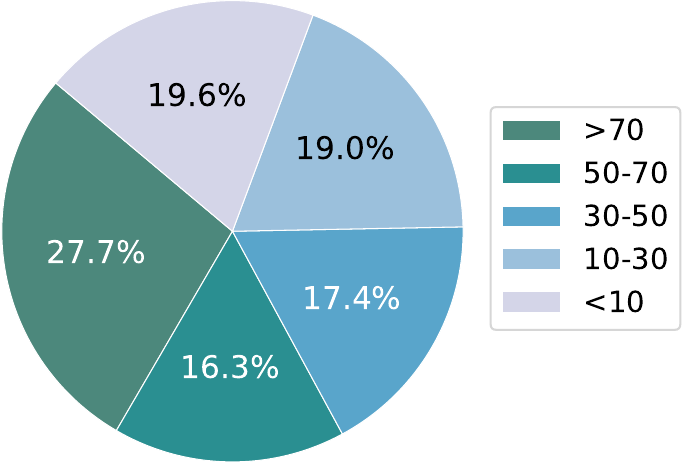}
    \caption{Distribution of GPU efficiency of the Triton operators in \benchone.}
    \label{fig:performance_distribution}
\end{figure}

To systematically evaluate the quality of the Triton operators in \benchone, we compute the GPU efficiency for each operator. 
Detailed methodology for calculating GPU efficiency can be found in Appendix~\ref{sec:appendix-performance_eval}.
Our statistics indicate an average GPU efficiency of $\mathbf{43.0}\%$, which reflects the overall reliability of the operators in \benchone. 
The distribution of efficiency scores is shown in Figure~\ref{fig:performance_distribution}. 
As shown in the figure, $19.6$\% of operators developed by professional Triton programmers have GPU performance below 10\%, which underscores the challenges in developing and optimizing Triton operators.

\subsection{Test Code Design}
\label{sec:benchone-testcodedesign}
In contrast to traditional CPU-language benchmarks~\cite{shypulalearning,evalplus,huang2024effibench,qiu2024efficient} that predominantly rely on scalar test inputs,\benchone is built around tensor-based test inputs. 
We employ PyTorch to generate random tensors as replacements for conventional test cases. 
Specifically, we leverage a prompt~\ref{prompt_test_code} to generate the corresponding test code for each operator. 
In the case of the multi-branch operators, the generated test code is designed to invoke every branch within the operator. 
Moreover, we rigorously debug all branches to guarantee test reliability. 
On average, we generate $3.6$ test branches per operator.

\subsection{Evaluation Metrics}
\label{sec:benchone-evalmetric}
In contrast to traditional code evaluations, which mainly emphasize accuracy~\cite{chen2021evaluating,austin2021program}, our \benchone introduces dedicated performance evaluations. Specifically, the systematic evaluation of Triton operators covers five key metrics:

\paragraph{\texttt{Similarity}} assesses text-level resemblance using \textsc{CodeBLEU} \cite{ren2020codebleu}. In our experiments, we assign equal weights of $0.25$ to N-gram, weighted N-gram, syntax, and dataflow components to ensure a balanced evaluation.
\paragraph{\texttt{Call~\&~Execution Accuracy}} assess whether the code can run without error and whether its input-output behavior is correct, respectively. 
\paragraph{\texttt{Speed Up}} measures the relative execution time improvement for correctly executed operators. Specifically, if \(t_{gen}\) and \(t_{ref}\) represent the running times of the generated and reference operators, respectively, then
\(
\texttt{SpeedUp}(gen) = \frac{t_{ref}}{t_{gen}}.
\)
\paragraph{\texttt{GPU Efficiency}} evaluates how effectively the generated operator utilizes GPU resources, following the operator quality rating in \S~\ref{sec:benchone-qualityrating}. For further details, please refer to Appendix~\ref{sec:appendix-performance_eval}.
\section{\benchtwo}
The real-world Triton operators introduced in \S~\ref{sec:git} primarily focus on highly frequent operations. 
As a complement, we propose \benchtwo, which aligns the Triton wrapper with interfaces of the PyTorch library \cite{paszke2019pytorch}. 
Together, \benchone and \benchtwo form a complementary evaluation framework.
The following sections elaborate on the data construction (\S~\ref{sec:benchtwo-construction}), data statistics (\S~\ref{sec:benchtwo-datastatistic}), test code and metrics (\S~\ref{sec:benchtwo-testandmetric}), and  benchmark comparisons (\S~\ref{sec:benchtwo-comparison}).

\subsection{Data Construction}
\label{sec:benchtwo-construction}
We construct \benchtwo by selecting PyTorch operators based on their usage frequency in real-world coding and then fusing them ( hereafter referred to simply as ``operators'' ). 
First, we select operators that require GPU interactions, ensuring alignment with Triton's scope.
Next, we sample $40$ high-frequency operators and $40$ low-frequency operators from the remaining pool. 
The frequency of each operator is determined by its usage probability in PyTorch-related code from The Stack V2~\cite{lozhkov2024starcoder} with those exceeding a predefined threshold $45$\% as common operators. 

Subsequently, we fuse these operators in various configurations: combinations of common operators, combinations of common and uncommon operators, and combinations of uncommon operators. 
All combinations are valid, as the outputs of preceding operators serve as appropriate inputs for subsequent ones. 
The final set includes $166$ operators, based on the latest (v2.6.0) version of the PyTorch library. 
Each operator is paired with its corresponding standard PyTorch call and document, while fused operators combine descriptions from all involved operators. 

\subsection{Data Statistic}
\label{sec:benchtwo-datastatistic}
The statistics of \benchtwo are presented in Table~\ref{tab:statis_torch}. 
Similar to \benchone, the operators are categorized into five difficulty levels ($\bf d1$ to $\bf d5$) using an LLM guided by prompt~\ref{prompt_diff}. 
These initial categorizations are then validated through manual review by two domain experts. 

We report the following statistics: (1)\textbf{torch-op\#} the average number of PyTorch operators, (2)\textbf{params\#} the average number of parameters, (3)\textbf{math\#}, the average token number of mathematical expressions, and (4)\textbf{description\#}, the average token count of the descriptions. 
These statistics generally increase with the operator difficulty, similar trend that aligns with the observations in \benchone.

\begin{table}[t!]
    \centering
    \resizebox{0.48\textwidth}{!}{
    \begin{tabular}{lccccccc}
    \toprule
    \multirow{2}{*}{\textbf{Difficulty}} & \multicolumn{4}{c}{\textbf{Torch-Align Operator}}    \\
    \cmidrule(lr){2-5}
     & \textbf{torch-op\#} & \textbf{params\#} & \textbf{math\#} & \textbf{description\#}       \\
    \midrule
    $\mathbf{d1}$ ($13.3$\%)    & $1.36$  & $2.82$  & $23.50$  & $50.41$   \\
    $\mathbf{d2}$ ($22.3$\%)    & $1.97$  & $3.78$  & $40.73$  & $61.19$   \\
    $\mathbf{d3}$ ($32.5$\%)    & $2.70$  & $4.91$  & $74.64$  & $67.89$   \\
    $\mathbf{d4}$ ($29.5$\%)    & $2.16$  & $5.24$  & $47.31$  & $71.02$   \\
    $\mathbf{d5}$ ($2.4$\%)     & $2.75$  & $2.75$  & $30.50$  & $88.50$   \\
    \bottomrule
    \end{tabular}
    }
    \caption{Statistics of \benchtwo.
    }
    \label{tab:statis_torch}
\end{table}

\subsection{Test Code and Metrics}
\label{sec:benchtwo-testandmetric}
The design of the test code in \benchtwo adheres to those of \benchone, employing randomly generated tensors for operator evaluation.
For correctness and performance assessment, we utilize \textbf{\texttt{Call Accuracy}}, \textbf{\texttt{Execution Accuracy}}, and \textbf{\texttt{Speed Up}}, whose computation methods are consistent with those used in \benchone.

\begin{table*}[ht!] \small
    \centering
    \resizebox{0.99\textwidth}{!}{
        \begin{tabular}{lcccccc}
        \toprule
        \textbf{Model}  & \textbf{Size} & \textbf{Similarity} & \textbf{\makecell{Call \\Accuracy}} & \textbf{\makecell{Execution \\Accuracy}} & \textbf{Speed Up} & \textbf{\makecell{GPU \\Efficiency}}  \\ 
        \midrule
            \multicolumn{7}{c}{\textit{Domain-Specific Models}} \\
        Qwen2.5-Coder      	& $7$B            &$9.19$~/~$14.54$               &$0.00$~/~$0.00$                            &$0.00$~/~$0.00$                      &$0.00$~/~$0.00$   &$0.00$~/~$0.00$   \\
        DeepSeek-Coder  	& $6.7$B          &$9.38$~/~$14.52$               &$0.00$~/~$0.00$                            &$0.00$~/~$0.00$                      &$0.00$~/~$0.00$   &$0.00$~/~$0.00$   \\
                
        Qwen2.5-Coder-sft	& $7$B            &$\mathbf{29.98}$~/~$25.96$     &$4.89$~/~$10.87$                         &$4.89$~/~$10.87$                   &$\mathbf{1.56}$~/~$\mathbf{1.22}$    &$\mathbf{51.71}$~/~$\mathbf{46.70}$     \\
        DeepSeek-Coder-sft & $6.7$B          &$25.52$~/~$\mathbf{30.34}$&$\mathbf{9.78}$~/~$\mathbf{11.96}$       &$\mathbf{9.87}$~/~$\mathbf{11.96}$ &$1.03$~/~$1.11$    &$47.68$~/~$42.26$     \\
        \midrule
                \multicolumn{7}{c}{\textit{General-Purpose Models}} \\
        GPT-4o                  & -             &$9.87$~/~$20.67$                 &$10.87$~/~$17.93$                        &$10.33$~/~$16.84$                    & $0.97$~/~$1.19$           &$48.80$~/~$\mathbf{53.33}$     \\
        Claude-3.5-Sonnet       & -             &$12.46$~/~$22.48$                &$10.33$~/~$20.11$                        & $9.79$~/~$19.57$                    & $0.90$~/~$\mathbf{1.54}$           &$\mathbf{59.31}$~/~$49.32$     \\
        Qwen2.5-72B             & $72$B           &$14.86$~/~$26.25$                &$11.41$~/~$16.85$                        &$10.87$~/~$16.31$                    & $0.96$~/~$1.19$   & $23.28$~/~$49.40$     \\
        DeepSeek-R1             &$685$B           &$\mathbf{19.96}$~/~$22.64$       &$13.59$~/~$22.83$                        &$13.05$~/~$22.83$                    & $\mathbf{1.11}$~/~$1.22$   & $44.83$~/~$46.70$     \\
        GPT-o1                  & -             &$16.58$~/~$\mathbf{29.70}$       &$\mathbf{15.22}$~/~$\mathbf{23.91}$      &$\mathbf{14.23}$~/~$\mathbf{23.91}$  &$0.92$~/~$1.14$   &$54.25$~/~$46.37$     \\
        \bottomrule
    \end{tabular}
    }
    \caption{Main results of \benchone across baseline models, where the left side of ``/'' represents the zero-shot results and the right side represents the one-shot results.}
    \label{tab:git_res}
\end{table*}

\subsection{Benchmark Comparison} 
\label{sec:benchtwo-comparison}
This section provides comparisons between \benchone and \benchtwo, which differ in key aspects and together provide a well-rounded evaluation. 
\paragraph{Source \& Distribution:} \benchone is collected from \textbf{GitHub} and reflects real-world programming demands with a concentration of frequently used operators, e.g., \texttt{Attention} at $20.0\%$, \texttt{MatMul} at $10.9\%$, \texttt{LayerNorm} at $6.5\%$, \texttt{SoftMax} at $3.8\%$. In contrast, \benchtwo, sourced from \textbf{PyTorch}, presents a more diverse operator set including both common and uncommon operators.
\paragraph{Instruction Generation:} \benchone combines \textbf{LLM generation} with \textbf{expert verification} while \benchtwo directly extracts instructions from \textbf{PyTorch documentation}. This difference underlines their complementary roles in probing different facets of the Triton generation. 
\paragraph{Evaluation Metrics:} Both benchmark channels assess \textbf{correctness} and \textbf{performance}. Additionally, \benchone incorporates a similarity-based assessment that offers direct comparisons with established implementations. 
In summary, the different designs of \benchone and \benchtwo enable a comprehensive and nuanced evaluation of Triton operator generation. 

\section{Experiments}
We conduct an extensive set of experiments on \benchall to rigorously evaluate the performance and capabilities of current LLMs.

\begin{table*}[ht!] \small
    \centering
    \begin{tabular}{lcccc}
        \toprule
        \textbf{Model} & \textbf{Size} & \textbf{Call Accuracy} & \textbf{Execution Accuracy} & \textbf{Speed Up} \\ 
        \midrule
        \multicolumn{5}{c}{\textit{Domain-Specific Models}} \\
        Qwen2.5-Coder           & $7$B        & $0.00$~/~$0.00$                     & $0.00$~/~$0.00$                       & $0.00$~/~$0.00$     \\
        DeepSeek-Coder          & $6.7$B      & $0.00$~/~$1.81$                     & $0.00$~/~$1.81$                       & $0.00$~/~$\mathbf{0.94}$     \\
        Qwen2.5-Coder-sft      & $7$B        & $17.47$~/~$16.27$                    & $17.47$~/~$15.67$                     & $\mathbf{0.98}$~/~$0.92$     \\
        DeepSeek-Coder-sft     & $6.7$B      & $\mathbf{19.28}$~/~$\mathbf{18.67}$  & $\mathbf{19.28}$~/~$\mathbf{16.26}$   & $0.91$~/~$0.85$     \\

        \midrule
        \multicolumn{5}{c}{\textit{General-Purpose Models}} \\

        GPT-4o                      & -      &$36.75$~/~$32.53$                     &$36.75$~/~$32.53$                      & $0.98$~/~$0.94$     \\
        Claude-3.5-Sonnet           & -      &$29.52$~/~$37.95$                     &$29.52$~/~$33.70$                      & $0.93$~/~$0.89$     \\
        Qwen2.5-72B                 & 72B    &$30.12$~/~$22.89$                     &$30.12$~/~$16.30$                      & $1.07$~/~$0.92$     \\
        DeepSeek-R1                 &$685$B  &$\mathbf{53.01}$~/~$\mathbf{45.78}$   &$\mathbf{53.01}$~/~$\mathbf{45.78}$    & $1.03$~/~$\mathbf{1.91}$     \\
        GPT-o1                      & -      &$32.53$~/~$43.37$                     &$32.53$~/~$43.37$                      & $\mathbf{1.21}$~/~$1.10$     \\

        \bottomrule
    \end{tabular}
    \caption{Main results of \benchtwo across baseline models, where the left side of ``/'' represents the zero-shot results and the right side represents the one-shot results.}
    \label{tab:torch_res}
\end{table*}

\subsection{Baselines and Setup}
\label{sec:baselines}
\benchall generally requires strong capabilities in code generation. Therefore, we select state-of-the-art LLMs that excel in programming tasks as baselines, including both specialized open-source models and general-purpose models.
For specialized open-source models, we choose \texttt{Qwen2.5-Coder-7B-Instruct} \cite{hui2024qwen2} and \texttt{deepseek-coder-6.7b-instruct} \cite{guo2024deepseek}.
For general-purpose models, we include \texttt{Claude-3.5-Sonnet-0620}\footnote{\href{https://www.anthropic.com/news/claude-3-5-sonnet}{https://www.anthropic.com/news/claude-3-5-sonnet}}, \texttt{GPT-4o-0806} \footnote{\href{https://openai.com/index/hello-gpt-4o}{https://openai.com/index/hello-gpt-4o}}, \texttt{qwen2.5-72B-Instruct} \cite{yang2024qwen2}, as well as the thought-driven models DeepSeek-R1 \cite{guo2025deepseek} and \texttt{GPT-o1-2024-12-17} \footnote{\href{https://openai.com/o1/}{https://openai.com/o1/}}.

In our experiments, all general-purpose models are deployed for direct inference. 
In contrast, domain-specific models undergo an additional supervised fine-tuning phase. 
Details of the training corpus can be found in \S~\ref{sec:appendix-trainingcorpus}.
For evaluation, we consider both zero-shot and one-shot scenarios. 
In the one-shot setting, a BM25-based retrieval method~\cite{bm25robertson2009} is utilized to select the most relevant prompt from the training corpus.

\subsection{Main results of \benchone}
\label{sec:main_res}
Table~\ref{tab:git_res} illustrates the performances of baselines on \benchone. 
It is evident that domain-specific models generally underperform compared to general-purpose models. 
However, fine-tuning $7$B domain-specific models with domain data significantly boosts accuracy. 
Qwen's accuracy rises from $0$ to $4.89$\%, and DeepSeek's from $0$ to $9.78$\% in zero-shot settings, with even more pronounced enhancements in one-shot settings due to the retrieval data from the same source as \benchone. 
The observed increase in \texttt{Speed Up} can be attributed to the relative simplicity of the correctly generated operators, which makes it easier for LLMs to produce efficient code.
The high \texttt{GPU efficiency} shares the similar reasons. 

General-purpose models, particularly DeepSeek-R1 and GPT-o1, excel across all metrics. 
Under one-shot conditions, DeepSeek-R1 achieves $22.83$\% in \texttt{Call and Execution Accuracy}, while GPT-o1 reaches $23.91$\%. 
The roughly $10$\% improvement from zero-shot to one-shot highlights the critical role of high-quality examples for Triton generation. 
Furthermore, the close alignment between \texttt{Call Accuracy} and \texttt{Execution Accuracy} indicates that only a few operators fail to produce correct results despite successfully invoked. 

DeepSeek-R1 also leads in GPU execution times, with an improvement of $1.11\times$ in zero-shot and $1.22\times$ in one-shot settings. 
While GPU efficiency is strong across most models, Qwen2.5-72B exhibits lower efficiency in zero-shot settings, likely due to a higher proportion of less efficient operators. 
Finally, \texttt{Similarity} provides corroborative insights, as its variations mirror trends observed in other metrics.

\subsection{Main Results of \benchtwo}
From Table~\ref{tab:torch_res}, we can observe that domain-specific models generally underperform general-purpose models. 
Nonetheless, fine-tuning with an $8k$ corpus considerably improves their performance. 
For instance, Qwen’s zero-shot \texttt{Execution Accuracy} rises from $0$ to $17.47$\%.
In contrast, its one-shot improvement ($15.67$\%) is slightly lower, likely due to the fact that the retrieved prompts and \benchtwo operators come from different sources (Github vs. Pytorch).

Among general-purpose models, DeepSeek-R1 demonstrates the strongest overall performance, achieving $53.01$\% \texttt{Call and Execution Accuracy} in the zero-shot setting. 
Although its accuracy drops by $7.23$\% in the one-shot setting, it still slightly surpasses GPT-o1. 
As for \texttt{Speed Up}, DeepSeek-R1 achieves the best performance of $1.91\times$ improvements.
Most performance improvements in successfully executed operators stem from operator fusion. Triton’s fused operators reduce redundant memory reads and writes compared to PyTorch, enhancing memory bandwidth utilization and boosting performance.

Overall, most models achieve better performance on \benchtwo than to \benchone, likely because \benchtwo features a more balanced distribution of operator difficulty, whereas \benchone is predominantly composed of higher-difficulty operators, namely, $\mathbf{d3}$ and $\mathbf{d4}$.

\section{Analysis}

In this section, we examine the distribution of correct and incorrect operators across difficulty levels ($\mathbf{d1}$–$\mathbf{d5}$) for the top-performing models, DeepSeek-R1 and GPT-o1, as shown in Figure~\ref{fig:barG} and Figure~\ref{fig:barT}. Additionally, we analyze the error patterns of incorrect operators and summarize the main challenges for each benchmark as detailed in Table~\ref{tab:git_res} and Table~\ref{tab:torch_res}.
The zero-shot and one-shot settings are annotated as $^0$ and $^1$ respectively.

\subsection{Challenges for \benchone}

\begin{figure}[t!]
    \centering
    \begin{subfigure}[t]{0.23\textwidth}
        \centering
        \includegraphics[width=\linewidth]{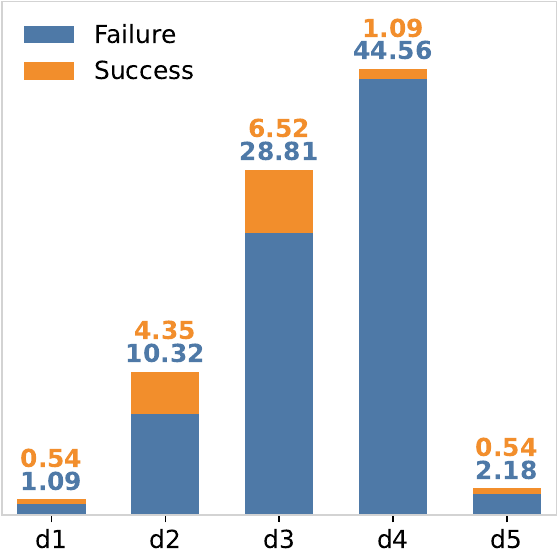}
        \caption{DeepSeek-R1$^0$ results.}
    \end{subfigure}
    \hspace{-0.008\textwidth}  
    \begin{subfigure}[t]{0.23\textwidth}
        \centering
        \includegraphics[width=\linewidth]{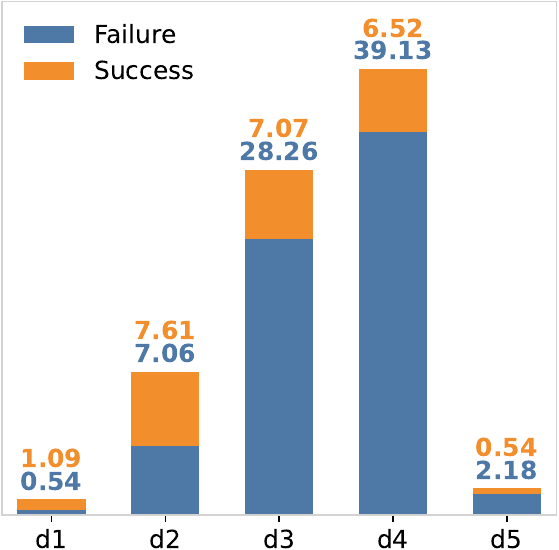}
        \caption{DeepSeek-R1$^1$ results.}
    \end{subfigure}
    
    \vspace{0.2cm}  

    \begin{subfigure}[t]{0.23\textwidth}
        \centering
        \includegraphics[width=\linewidth]{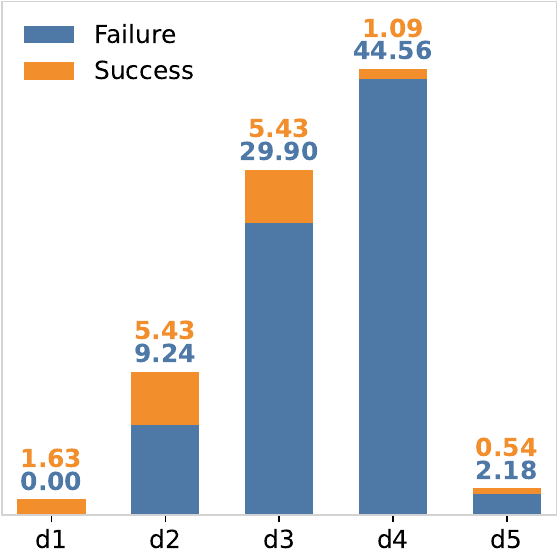}
        \caption{GPT-o1$^0$ results.}
    \end{subfigure}
    \hspace{-0.008\textwidth}  
    \begin{subfigure}[t]{0.23\textwidth}
        \centering
        \includegraphics[width=\linewidth]{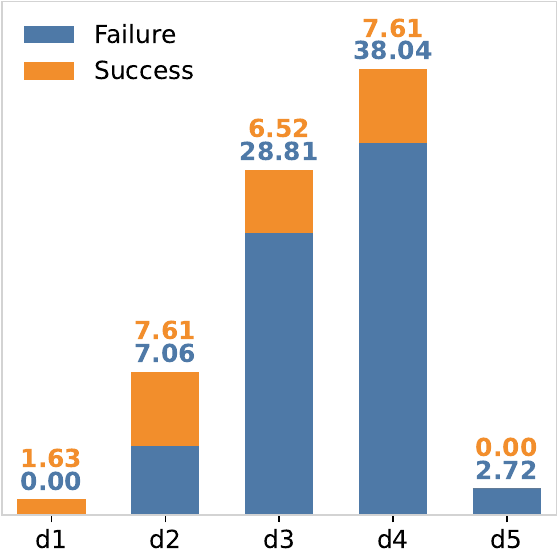}
        \caption{GPT-o1$^1$ results.}
    \end{subfigure}

    \caption{Execution results distribution across difficulty levels in \benchone.}
    \label{fig:barG}
\end{figure}

Figure~\ref{fig:barG} clearly shows that most operators are generated incorrectly. 
Both DeepSeek-R1 and GPT-o1 exhibit similar trends, with DeepSeek-R1 outperforming GPT-o1.
Notably, when moving from the zero-shot to the one-shot setting, both models achieve significant improvements on $\mathbf{d4}$.
These improvements may stem from the prevalence of \texttt{Attention} and \texttt{Softmax} operators in $\mathbf{d4}$, enabling models to leverage similar examples. 
In contrast, the simpler operators in $\mathbf{d2}$ and $\mathbf{d3}$ show only limited gains in the one-shot setting, likely due to the smaller, more idiosyncratic nature of these datasets that leads to lower retrieval similarity.

For the incorrectly written operators, we classify the $16$ error types into $4$ major categories, detailed in Appendix~\ref{app:error_catgrz}
which is presented in Table~\ref{tab:err_G}.
Note that only compiler-reported errors were considered.
The results show that, compared to the zero-shot setting, both DeepSeek-R1 and GPT-o1 in the one-shot setting demonstrate a significant increase in \texttt{Syntax} and \texttt{Name\&Ref} errors but a reduction in \texttt{Attr\&Type} and \texttt{Run\&Logc} errors. 
This trend suggests that the training corpus may provide helpful guidance on logical structure and Triton specifications, thus enhancing overall accuracy.
Furthermore, error sensitivity differs between models: DeepSeek-R1 is less susceptible to syntax errors, whereas GPT-o1 handles logical errors better.

\subsection{Challenges for \benchtwo}
\begin{figure}[t!]
    \centering
    \begin{subfigure}[t]{0.23\textwidth}
        \centering
        \includegraphics[width=\linewidth]{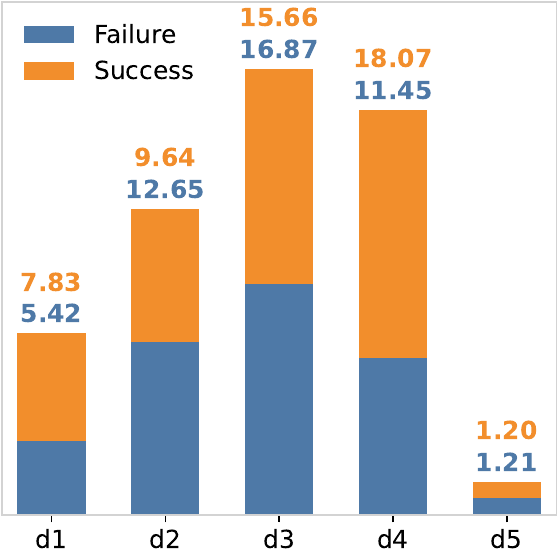}
        \caption{DeepSeek-R1$^0$ results.}
    \end{subfigure}
    \hspace{-0.008\textwidth}  
    \begin{subfigure}[t]{0.23\textwidth}
        \centering
        \includegraphics[width=\linewidth]{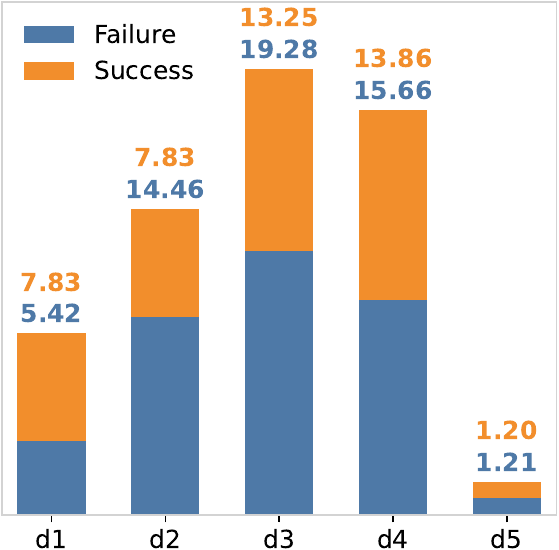}
        \caption{DeepSeek-R1$^1$ results.}
    \end{subfigure}
    
    \vspace{0.2cm}  

    \begin{subfigure}[t]{0.23\textwidth}
        \centering
        \includegraphics[width=\linewidth]{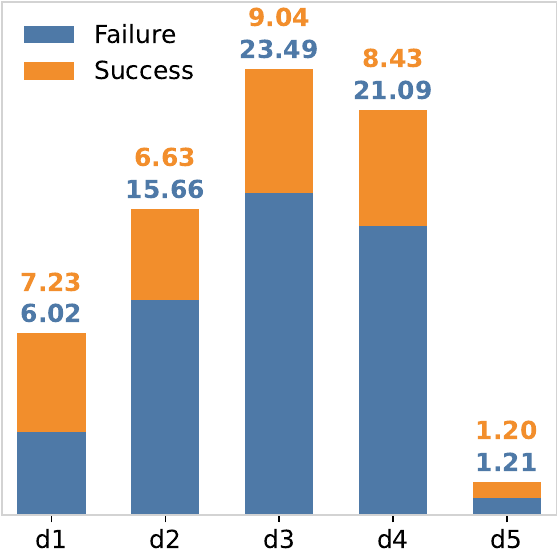}
        \caption{GPT-o1$^0$ results.}
    \end{subfigure}
    \hspace{-0.008\textwidth}  
    \begin{subfigure}[t]{0.23\textwidth}
        \centering
        \includegraphics[width=\linewidth]{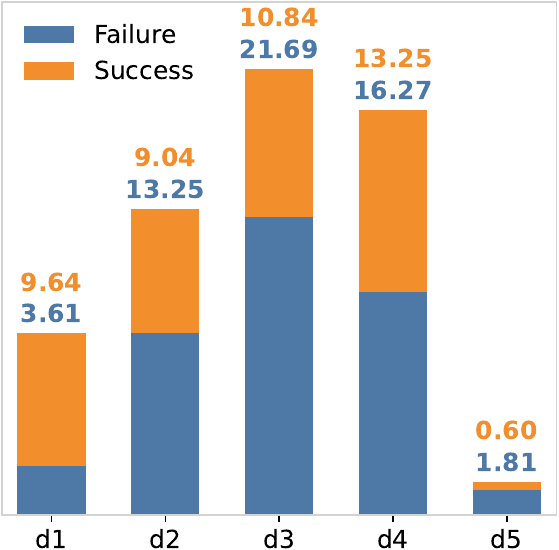}
        \caption{GPT-o1$^1$ results.}
    \end{subfigure}

    \caption{Execution results distribution across difficulty levels in \benchtwo.}
    \label{fig:barT}
\end{figure}

The execution results of \benchtwo (Figure~\ref{fig:barT}) show the percentages of correctly generated operators.
we can observe that DeepSeek-R1 generated more correct than incorrect operators, which proves the point that the difficulty distributions in \benchtwo are smoother than \benchone. 

However, while DeepSeek-R1's performance declines for difficulty $\mathbf{d2}$-$\mathbf{d4}$ in the one-shot setting, GPT-o1 shows improved accuracy on these subsets. 
This finding indicates that GPT-o1 might be more adept at logical reasoning for Triton generation tasks, allowing it to efficiently use the provided sample. The differing trends also imply that sample operators affect models in diverse ways.

\begin{table}[t]
    \centering
    \resizebox{0.48\textwidth}{!}{
    \begin{tabular}{l cccc}
        \toprule        
        \textbf{Model}  &\textbf{Syntax}    &\textbf{Attr\&Type}     &\textbf{Name\&Ref}  &\textbf{Run\&Logc}      \\
        \midrule
        DeepSeek-R1$^0$            &$1.64$               &$42.62$                  &$16.39$              &$39.34$   \\
        DeepSeek-R1$^1$            &$9.27$               &$33.11$                  &$35.76$              &$21.85$   \\
        GPT-o1$^0$            &$10.3$               &$38.18$                  &$28.48$              &$23.03$   \\ 
        GPT-o1$^1$            &$20.83$              &$24.31$                  &$43.06$              &$11.81$   \\        
        \bottomrule
    \end{tabular}}
    \caption{Error statistics of execution failures in \benchone.}
    \label{tab:err_G}
\end{table}

For execution error statistics in \benchtwo (Table~\ref{tab:err_T}), DeepSeek-R1 notably avoids \texttt{Syntax} errors entirely, while GPT-o1 maintains a high rate of such errors. 
Under the one-shot setting, DeepSeek-R1 shows a rise in \texttt{Attr\&Type} and \texttt{Name\&Ref} errors alongside a decline in \texttt{Run\&Logc} Errors. 
Conversely, GPT-o1 experiences a significant increase in \texttt{Name\&Ref} errors with a notable drop in \texttt{Run\&Logc} errors. 
Comparing \benchone and \benchtwo, the one-shot setting consistently reduces \texttt{Run\&Logc} errors. 
These variations in error patterns likely stem from the mixed influence of useful and irrelevant information in the provided samples.

\begin{table}[t]
    \centering
    \resizebox{0.48\textwidth}{!}{
    \begin{tabular}{l cccc}
        \toprule        
        \textbf{Model}  &\textbf{Syntax}    &\textbf{Attr\&Type}     &\textbf{Name\&Ref}  &\textbf{Run\&Logc}      \\
        \midrule
        DeepSeek-R1$^0$        &$0.00$        &$31.96$       &$14.43$       &$53.61$   \\
        DeepSeek-R1$^1$        &$0.00$        &$36.79$       &$20.75$       &$42.45$   \\
        GPT-o1$^0$        &$24.06$       &$26.32$       &$7.52$        &$42.11$   \\ 
        GPT-o1$^1$        &$25.25$       &$25.25$       &$22.22$       &$27.27$   \\        
        \bottomrule
    \end{tabular}}
    \caption{Error statistics of execution failures in \benchtwo.}
    \label{tab:err_T}
\end{table}

\section{Conclusion}

In this work, we present \benchall, a dual-channel benchmark specifically designed for evaluating LLMs' generation for Triton operators. \benchone integrates real-world Triton operator samples from open repositories, while \benchtwo introduces complementary tasks that align with PyTorch interfaces. Our evaluation framework addresses both functional accuracy and the performance on NVIDIA GPUs. 
We also conduct extensive experiments and detailed analysis on our benchmark, and find that current LLMs struggle to generate high-quality Triton operators, underscoring the necessity for further advancement in generating accurate as well as performance-aware Triton code. 
We anticipate \benchall will serve as an essential framework for advancing automated operator generation for Triton.
\section*{Limitations}
The primary limitation of this study is that the evaluations of \benchall were conducted exclusively on the NVIDIA A100 GPU, as it is widely adopted in industry and research applications. In future work, we plan to expand the evaluation to include a broader range of hardware architectures for more comprehensive performance insights.

\bibliography{anthology,custom}
\bibliographystyle{acl_natbib}

\appendix
\label{sec:appendix}

\section{Training Corpus}
\label{sec:appendix-trainingcorpus}
The training corpus for supervised fine-tuning comprises two distinct components: real-world data sourced from GitHub and synthetically generated data produced through compiler operations.

The real-world data component incorporates Triton code extracted from GitHub repositories, which undergoes basic cleaning procedures as outlined in prompt~\ref{prompt_filter}, undergoes a debugging process that is less rigorous than the methodology applied to \benchone.
To prevent potential data leakage and ensure benchmark integrity, we systematically eliminate samples exhibiting high similarity to \benchone entries using the \textsc{CodeBertScore} similarity metric~\citep{codebertscore2023}.

The synthetic data component is generated using Ninetoothed\footnote{\href{https://github.com/InfiniTensor/ninetoothed}{https://github.com/InfiniTensor/ninetoothed}}, a domain-specific language built upon Triton that offers enhanced abstraction capabilities. This framework facilitates the automated synthesis of valid Triton code through the processing of well-formed expressions.
Each part of data containing $4K$ samples. This combined corpus serves as the foundational training dataset for experimental models in one-shot learning settings. For all experiments, the fine-tuning process is carried out over $3$ epochs with a learning rate of $5e-5$. 

\section{Operator Performance Evaluation}
\label{sec:appendix-performance_eval}
\begin{figure}
    \centering
    \includegraphics[width=7.5cm]{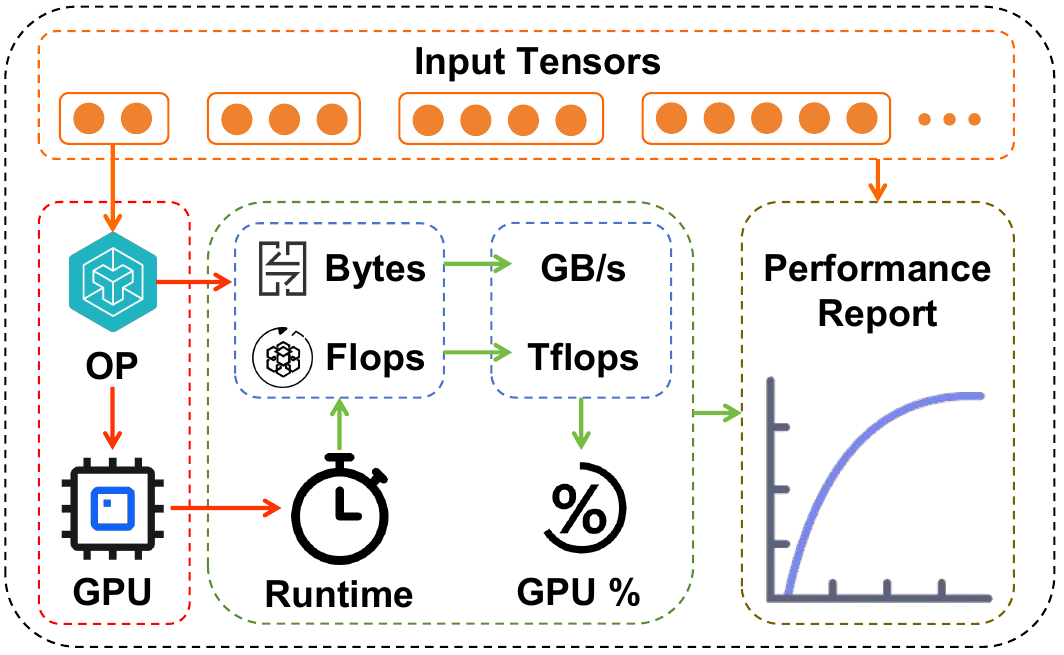}
    \caption{The workflow of operator performance evaluation}
    \label{fig:performance_evaluation_metrics}
\end{figure}

For operator performance evaluation, we refer primarily to the official examples provided by Triton\footnote{\href{https://triton-lang.org/main/getting-started/tutorials/}{https://triton-lang.org/main/getting-started/tutorials/}}. 
We provide evaluation scripts for each operator in \benchone.
Figure~\ref{fig:performance_evaluation_metrics} illustrates the workflow of our operator performance evaluation.

First, we define a set of tensors with increasing dimensions based on the characteristics of the operator. 
Next, each tensor is sequentially fed into the operator for execution. 
During each execution, we use the expert annotations for each operator to determine the total memory bandwidth (Bytes) and the total number of floating-point operations (Flops) based on the input tensors.
More importantly, we use the \texttt{triton.testing.do\_bench} method from the official Triton library\footnote{\href{https://triton-lang.org/main/python-api/generated/triton.testing.do_bench.html}{https://triton-lang.org/main/pythonapi/generated/triton.testing.do\_bench.html}} to measure the operator’s execution time on the GPU. 
Specifically, we gradually increase the warm-up time and repetition time until the measured execution time stabilized, which means that most operators are run hundreds of thousands of times to ensure that the running time is measured accurately.
After obtaining the execution time, we calculate the operator’s performance metrics by dividing the total memory bandwidth and the total floating-point operations by the execution time to obtain throughput in GB/s and Tflops, respectively. 
We then calculate the GPU efficiency by calculating the ratio of the measured performance metrics (GB/s and Tflops) to the theoretical maximum performance of the NVIDIA A100 Tensor Core GPU.
Repetition of the above process for tensors of increasing sizes obtains the performance metrics for each execution, which collectively form the operator performance report.
We adopt the peak GPU efficiency from the performance report as the final measure of the operator's quality.

\begin{figure*}
    \centering
    \includegraphics[width=16cm]{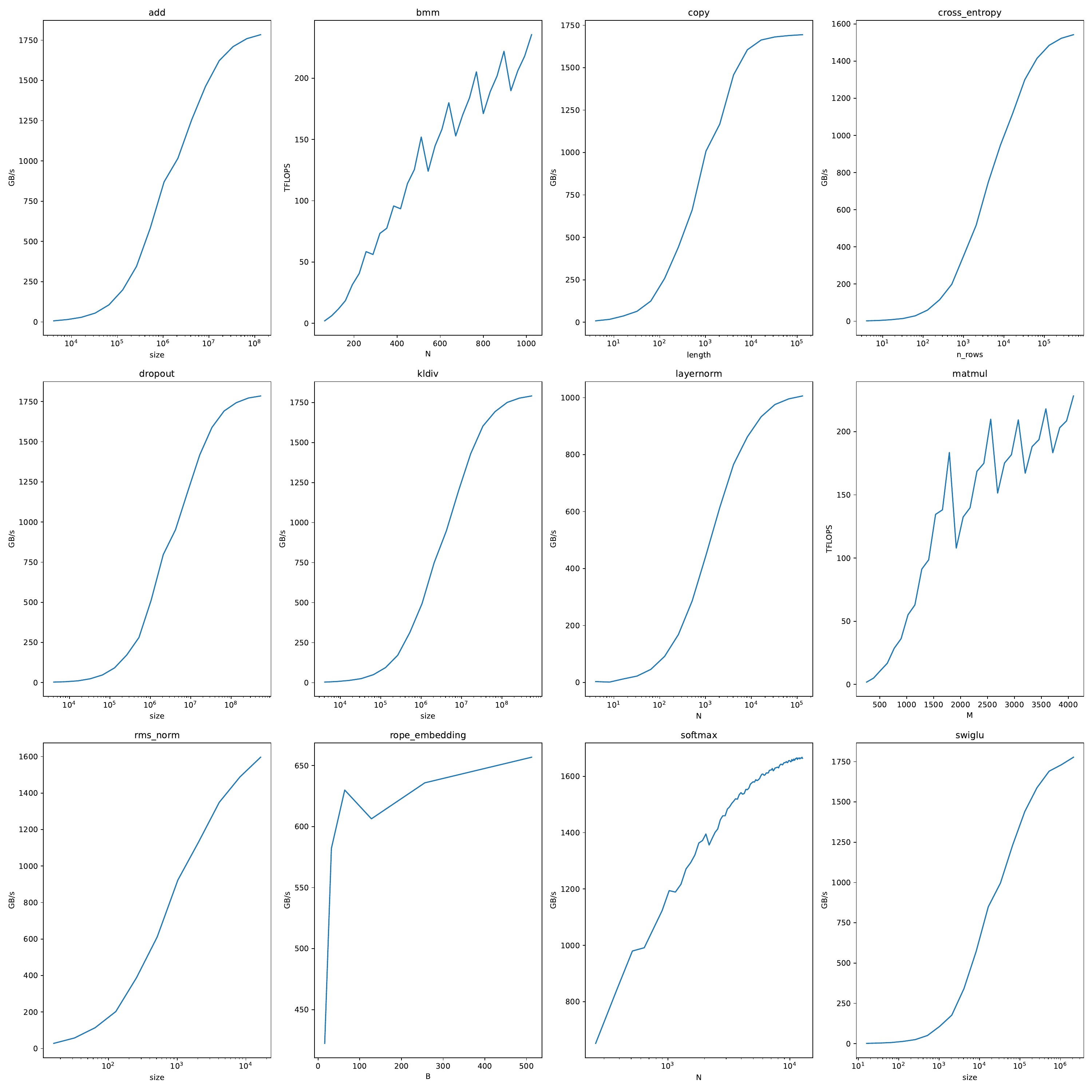}
    \caption{Performance Curves of Common Operators}
    \label{fig:common_operators_curves}
\end{figure*}

By following the evaluation workflow described above, we generate a detailed performance report for each operator in \benchone. Figure~\ref{fig:common_operators_curves} illustrates the performance curves of several common operators. As the input dimensions increase, as can be seen from the figure, the GB/s or Tflops of the operators show an upward trend, eventually stabilizing. 
This suggests that the performance of the operator reaches a bottleneck beyond a certain scale, and further increases in input size result in diminishing returns in performance, aligning with the expected trend of operator performance.

\section{Error Categories}
\label{app:error_catgrz}
We provide the error type statistics of failure operators in \benchall. A total of 16 error types are identified in the integrated Call and Execution error results. For convenience in presentation, we categorize them into four main groups: Syntax Errors: including SyntaxError and IndentationError; Attrb\&Type Errors: including AttributeError, TypeError, and NotImplementedError; Name\&Ref Errors: including NameError, KeyError, IndexError, ModuleNotFoundError, and ImportError; Run\&Logc Errors: including ValueError, ZeroDivisionError, RuntimeError, RecursionError, AssertionError, CompilationError, and ResultsError. ResultsError refers to the inconsistency between the execution results of the reference operator and the generated operator.

\section{Prompts}
Here are the four prompts we use in our work: Filtering Prompt, Instruction Prompt, Difficulty Prompt, and Test Code Prompt. Specifically, the first is used to extract Triton-related code from crawled code files; the second instructs the large model to generate corresponding instructions based on Triton code; the third prompts the large model to score the difficulty of Triton operators according to the standards we proposed; and the last asks the large model to generate test code.
\onecolumn

\begin{tcolorbox}[colframe=gray!80!black, colback=gray!10!white, title=Filtering Prompt]
\small
{\{code\}}\\
\justifying
Please help me select \textbf{all} triton kernel functions decorated with \texttt{@triton.jit} and all code that calls these kernels, while only keeping the necessary imports (e.g., triton, torch) and the calling functions. \\
\hdashrule[0.5ex]{15cm}{0.1pt}{1mm}
\justifying \\
Note 1: Retain necessary comments related to the Triton code. Code can be optimized, but do not remove all kernel code and its corresponding calls just for brevity. \\
\hdashrule[0.5ex]{15cm}{0.1pt}{1mm}
\justifying \\
Note 2: If the triton kernel is decorated with a custom or third-party decorator other than triton, discard that kernel. \\
\hdashrule[0.5ex]{15cm}{0.1pt}{1mm}
\justifying \\
Note 3: If \texttt{@triton.jit} appears as a \textbf{string} in the code or is nested within a function body, then discard it. \\
\hdashrule[0.5ex]{15cm}{0.1pt}{1mm}
\justifying \\
Note 4: If there are multiple triton kernel functions decorated with \texttt{@triton.jit} and their calling wrapper functions, retain all of them, not just a subset.

1) Extract all triton operators (kernel functions decorated with \texttt{@triton.jit} and their calling functions) and output them in python code format. If no triton kernel function is found, discard it.

2) Provide a concise English description of each extracted operator (including both kernel and calling code) in the form of a python dictionary: \texttt{"description": "Use triton language to..."}
\label{prompt_filter}
\end{tcolorbox}

\begin{tcolorbox}[colframe=gray!80!black, colback=gray!10!white, title=Instruction Prompt]
\small
\texttt{\{code\}}\\ \justifying 
Based on the above Triton operator code, generate a detailed description so that the large model can accurately reproduce the corresponding kernel and wrapper function. \\
Be clear about the logic and main functionality of the operator, specify the function name, inputs, and outputs, and describe any public variables clearly. \\
Try to describe the function's code implementation. Ensure that the large model can reproduce the corresponding function and parameter code based on these instructions. \\
Note that the output should maintain correct python syntax. \\
\label{prompt_instru}
\end{tcolorbox}

\begin{tcolorbox}[colframe=gray!80!black, colback=gray!10!white, title=Test Code Prompt]
\small
\texttt{\{code\}}\\ \justifying
Write a test code in Python for the above code. Ensure that all branch tests are in a single function starting with \texttt{``test\_''}, with no parameters.\\
\hdashrule[0.5ex]{15cm}{0.1pt}{1mm}
\justifying
Note 1: Particular attention should be paid to the fact that tensor parameters are of GPU type. \\
\hdashrule[0.5ex]{15cm}{0.1pt}{1mm}
\justifying \\
Note 2: Try to limit the number of branches to no more than 4. \\
\hdashrule[0.5ex]{15cm}{0.1pt}{1mm}
\justifying \\
Note 3: In branch tests, avoid modifying parameters that are later in the argument list with default values (especially if they have out parameters, do not assign them). \\
\hdashrule[0.5ex]{15cm}{0.1pt}{1mm}
\justifying \\
Note 4: Store the results of all branch calculations in a dictionary, where the dictionary key is \texttt{"test\_case\_n"}, with \texttt{n} representing the test case number.\\
\hdashrule[0.5ex]{15cm}{0.1pt}{1mm}
\justifying \\
Note 5: Ensure that the import paths match exactly as described in the operator documentation to maintain accuracy.\\
\hdashrule[0.5ex]{15cm}{0.1pt}{1mm}
\justifying \\
Note 6: The code should run directly, without \texttt{if \_\_name\_\_ == "\_\_main\_\_"}. \\
\label{prompt_test_code}

\end{tcolorbox}
\begin{tcolorbox}[colframe=gray!80!black, colback=gray!10!white, title=Difficulty Prompt]
\small
\texttt{\{code\}}\\ \justifying 
Please evaluate the complexity of the code in the following two aspects based on the requirements of the Triton operator and score it from simple to complex on a scale from 1 to 5: \\
1) Memory layout complexity: Analyze the memory access pattern, including memory tiling, array transposition, address alignment, cache utilization, and the number of global memory accesses. \\
2) Computation scheduling complexity: Examine instruction-level parallelism, computation-memory pipeline, thread block design, inter-thread communication, thread branch divergence, and hardware resource utilization. \\
The final score is the ceiling of the average score from both aspects, and only one complexity score is output in [ ].
\label{prompt_diff}
\end{tcolorbox}

\end{document}